\documentclass{article}

\PassOptionsToPackage{numbers,compress}{natbib}

\usepackage[preprint]{neurips_2026}

\usepackage[utf8]{inputenc} 
\usepackage[T1]{fontenc}    
\usepackage{hyperref}       
\usepackage{url}            
\usepackage{booktabs}       
\usepackage{amsfonts}       
\usepackage{amsmath}        
\usepackage{amssymb}        
\usepackage{nicefrac}       
\usepackage{microtype}      
\usepackage[table]{xcolor}  
\usepackage{comment}
\usepackage{algorithm}
\usepackage{algpseudocode}
\usepackage{multirow}
\usepackage{graphicx}
\usepackage{wrapfig}        
\usepackage{caption}        
\usepackage{titlesec}       

\titlespacing*{\section}      {0pt}{1.0ex plus 0.3ex minus 0.2ex}{0.6ex plus 0.2ex}
\titlespacing*{\subsection}   {0pt}{0.8ex plus 0.2ex minus 0.2ex}{0.4ex plus 0.2ex}
\titlespacing*{\subsubsection}{0pt}{0.6ex plus 0.2ex minus 0.2ex}{0.3ex plus 0.2ex}
\titlespacing*{\paragraph}    {0pt}{0.4ex plus 0.2ex minus 0.1ex}{0.8em}
\selectfont

\title{Can Graphs Help Vision SSMs See Better?}

\author{%
  \textbf{Dhruv Parikh}\textmd{\textsuperscript{1}}, \textbf{Anvitha Ramachandran}\textmd{\textsuperscript{1}}, \textbf{Haoyang Fan}\textmd{\textsuperscript{1}} \\
  \textbf{Mustafa Munir}\textmd{\textsuperscript{2}}, \textbf{Rajgopal Kannan}\textmd{\textsuperscript{3}}, \textbf{Viktor Prasanna}\textmd{\textsuperscript{1}} \\[0.4em]
  \textsuperscript{1}USC \quad \textsuperscript{2}UT Austin \quad \textsuperscript{3}DEVCOM ARL Army Research Office, USA \\[0.2em]
  \texttt{\{dhruvash, alramach, haoyangf, prasanna\}@usc.edu},\ \\ \texttt{mmunir@utexas.edu}, 
  \texttt{rajgopal.kannan.civ@army.mil}
}

\begin{document}

\maketitle
\vspace{-3mm}

\begin{abstract}
Vision state space models inherit the efficiency and long-range modeling ability of Mamba-style selective scans. However, their performance depends critically on the representation of two-dimensional visual features as one-dimensional token sequences. Existing scan operators range from predefined geometric traversals to dynamic coordinate-based samplers that reroute tokens through predicted offsets and interpolation. While effective, these mechanisms primarily adapt paths or sampling locations, rather than explicitly modeling which local patches should exchange information before global state-space mixing. This motivates a simple question: \emph{can graphs help vision state space models see better?} We introduce \textbf{GraphScan}, a graph-induced dynamic scanning operator for Vision SSMs. For each token, GraphScan constructs a spatially bounded local graph, learns feature-conditioned affinities with relative positional bias, and produces the output token by one-step message passing over its semantic neighborhood. The resulting tokens are locally grounded before being processed by the selective SSM for global aggregation. GraphScan preserves token count and linear scaling in image size, while replacing coordinate-conditioned interpolation with feature-conditioned semantic routing. Integrated into a hierarchical backbone, \textbf{GraphScan-Mamba} achieves state-of-the-art performance among Vision SSMs across image classification, object detection, instance segmentation, and semantic segmentation, with modest computational overhead. Our analysis further shows that GraphScan induces interpretable displacement fields over the token lattice, providing a semantic and spatially grounded view of dynamic scanning. These results suggest that future Vision SSMs should treat scanning not merely as geometric serialization, but as learned local semantic routing before global state-space modeling.
\end{abstract}
\section{Introduction}
\label{sec:intro}

The design of visual backbones has repeatedly been shaped by how an image is represented before it is processed. Convolutional networks preserve the image as a regular grid \cite{lenet,alexnet,vgg,googlenet,resnet,densenet,efficientnet,regnet,convnext,convnextv2,repvit,unireplknet,slak,inceptionnext} while Vision Transformers recast it as a sequence of patches with self-attention for long-range interaction \cite{transformer,vit,deit,swin,pvt,pvtv2,twins,cswin,focal,nat,dat,biformer,maxvit,coatnet,vitadapter,internimage}; all-MLP and graph-based variants further show this representation choice is broader than convolution or attention alone \cite{mlpmixer,resmlp,gmlp,vig}. More recently, structured state space models (SSMs) have emerged as efficient sequence mixers with long-range modeling and linear scaling \cite{s4,s4d,s5,h3,hyena,mamba,mamba2,mamba3}, raising a basic question for vision: if the model is fundamentally a sequence processor, what is the right sequence representation for an image?

Vision Mamba backbones expose this question sharply: Mamba-style selective SSMs offer linear-complexity content-dependent sequence modeling \cite{mamba}, but applying them to two-dimensional images requires serializing a feature map into a token sequence---making the scan operator the architectural interface between image geometry and state-space computation. Early Vision Mamba models use bidirectional or two-dimensional selective scans \cite{vim,vmamba}, and subsequent work has explored continuous, local-window, atrous, multi-scale, grouped, multi-dimensional, and fractal traversal patterns \cite{plainmamba,localmamba,efficientvmamba,msvmamba,groupmamba,mamband,fractalmamba}, establishing scanning as a central design lever in Vision SSMs.

\begin{figure}[t]
    \centering
    \includegraphics[width=0.9\textwidth]{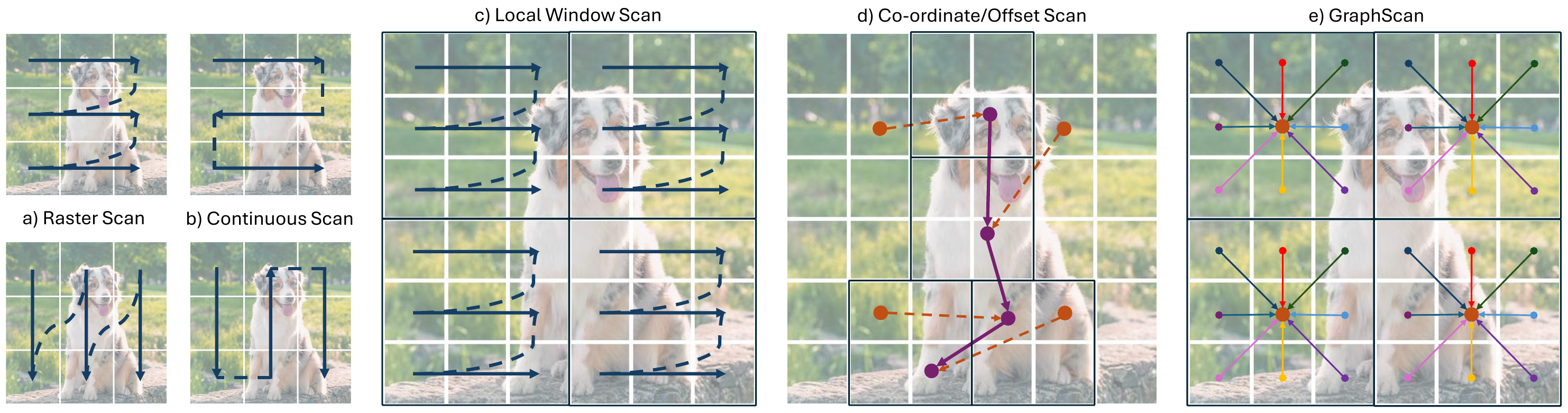}
    \caption{\textbf{Comparison of scanning paradigms for Vision SSMs.} Raster and continuous scans expose the SSM to a fixed 1D ordering of the 2D lattice; local-window scans restrict it to bounded predefined regions; coordinate/offset scans introduce input dependence by predicting sampling locations and rerouting through geometric displacement. \textbf{GraphScan} instead keeps token slots fixed and constructs a local semantic graph around each token, routing nearby patches through learned inter-patch affinities before global selective scan---scanning as local semantic routing rather than path or coordinate selection.}
    \label{fig:paradigm_comparison}
\end{figure}

A second wave of methods asks whether scanning can be made adaptive.
Some approaches learn spatial partitions or deformable scan paths, while others predict offsets, objectness scores, dynamic state aggregation patterns, or task-specific scan orders \cite{quadmamba,damamba,defmamba,objectness_scan,pvmamba,asm_unet,mambamatcher}.
Other works attack the problem from the recurrence side, for example by introducing structure-aware state fusion, non-causal SSD formulations, or hybrid Mamba-Transformer designs \cite{spatialmamba,vssd,mambavision}.
Together, these works reveal two complementary facts: visual SSMs need strong global sequence modeling, but they also need a better way to preserve the local and semantic structure of images before global propagation. Fig.~\ref{fig:paradigm_comparison} summarizes these scan paradigms and highlights the semantic-routing axis introduced by GraphScan.

\begin{table}[t]
\centering
\footnotesize
\setlength{\tabcolsep}{4pt}
\caption{\footnotesize Families of scanning mechanisms in Vision SSMs. \emph{Path}: how the 1D order through the lattice is determined. \emph{Adaptive}: whether the operator is input-dependent. \emph{Local mixing}: whether the operator explicitly aggregates a bounded spatial neighborhood before the SSM. GraphScan occupies the only cell that is simultaneously local, input-adaptive, and semantic.}
\label{tab:scan_families}
\begin{tabular}{lcccl}
\toprule
Family & Path & Adaptive & Local mixing & Representative works \\
\midrule
Fixed multi-direction         & predefined & --                          & implicit              & \cite{vim,vmamba,plainmamba,efficientvmamba,mamband,fractalmamba} \\
Local windowed                & predefined & --                          & fixed window          & \cite{localmamba,msvmamba,groupmamba} \\
Coordinate-offset             & raster     & learned offsets             & 4-corner bilinear     & \cite{damamba,defmamba,quadmamba} \\
Score / sort                  & permuted   & learned scalar              & --                    & \cite{objectness_scan,asm_unet,mambamatcher,pvmamba} \\
Recurrence-side               & --         & varies                      & varies                & \cite{spatialmamba,vssd,mambavision} \\
\textbf{Graph routing (ours)} & raster     & \textbf{learned affinities} & \textbf{semantic window} & \textbf{GraphScan} \\
\bottomrule
\end{tabular}
\end{table}

Despite this progress, most adaptive scans remain primarily geometric: coordinate-offset methods sample at predicted positions through bilinear interpolation \cite{damamba,defmamba}, score- and sort-based methods reorder tokens without aggregating local semantic neighborhoods \cite{objectness_scan,mambamatcher}, and recurrence-side methods modify the SSM mixer itself rather than the visual tokens entering it \cite{spatialmamba,vssd}. This leaves open a simple possibility: before asking an SSM to propagate information globally, can we first make each token locally and semantically aware?

We answer this question with \textbf{GraphScan}, a graph-induced dynamic scanning operator for Vision Mamba. Instead of predicting a continuous sampling coordinate, GraphScan constructs a spatially bounded local graph around each token and learns feature-conditioned affinities with relative positional bias. 
Each output token is produced by one-step message passing over this semantic neighborhood and then passed to the selective SSM for global aggregation. 
GraphScan therefore does not replace the SSM with a graph neural network  \cite{gcn,gat,vig}; rather, it uses graph-based routing as a lightweight scan-time operator that prepares locally grounded tokens for the global sequence mixer.

This gives GraphScan a distinct role among existing scan mechanisms: where fixed scans choose a path, deformable scans choose coordinates, and score-based scans choose an ordering, GraphScan chooses a local semantic neighborhood. The distinction matters because the selective SSM compresses visual context along a one-dimensional processing path; if the tokens entering it already encode spatially nearby and semantically relevant neighbors, the global mixer receives a better-conditioned sequence without changing its recurrence. GraphScan is therefore an input-side complement to recurrence-side advances such as SSD, non-causal SSD, and newer Mamba variants \cite{mamba2,vssd,mamba3}.

We instantiate GraphScan in a hierarchical Vision Mamba backbone, \textbf{GraphScan-Mamba}, on the Mamba-1/S6 selective-scan interface used by the dominant scan-mechanism literature \cite{vim,vmamba,plainmamba,localmamba,efficientvmamba,msvmamba,spatialmamba,damamba,defmamba}, enabling direct comparison with fixed, local, multi-scale, deformable, and dynamic scan baselines. GraphScan is orthogonal to the choice of SSM core: it changes the tokens presented to the mixer, not the mixer dynamics.

We evaluate GraphScan-Mamba under standard vision-backbone protocols (ImageNet-1K classification, COCO detection/instance segmentation with Mask R-CNN, ADE20K segmentation with UperNet \cite{imagenet,coco,maskrcnn,ade20k,upernet}), where it achieves state-of-the-art performance among Vision SSM backbones at comparable parameter and FLOP budgets. Ablations isolate the role of semantic routing, neighborhood size, relative positional bias, and pre-SSM insertion, and visualizations confirm that GraphScan induces interpretable displacement fields over the token lattice. Together, these results suggest a new design principle: scanning should not be treated as geometric serialization but as learned semantic routing before global state-space modeling.
\section{Related Work}
\label{sec:related}

\subsection{Vision Backbones and SSMs}
\label{sec:rw_backbones_ssm}

Vision backbones span convolutional \cite{convnext,convnextv2,inceptionnext}, Transformer \cite{vit,deit,swin,nat,biformer}, MLP-mixing \cite{mlpmixer}, and graph-based \cite{vig,pvg} families; we refer to App.~\ref{app:related_extended} for a fuller treatment. State-space sequence models provide efficient long-range mixing with near-linear scaling \cite{s4,s4d,s5,h3,hyena}, and Mamba introduced selective state spaces with input-dependent recurrence parameters \cite{mamba}, later extended through structured state space duality and recurrence-level improvements \cite{mamba2,mamba3}. Our work addresses a complementary question: how should a two-dimensional visual feature map be routed before it is processed by a one-dimensional selective scan?

\subsection{Vision Mamba Scans and Dynamic Routing}
\label{sec:rw_vssm_scans}

Vision Mamba models convert feature maps to token sequences using fixed scan patterns---bidirectional, two-dimensional, continuous, local-window, atrous, multi-scale, grouped, or space-filling \cite{vim,vmamba,plainmamba,localmamba,efficientvmamba,msvmamba,groupmamba,mamband,fractalmamba}. Adaptive variants reroute tokens before the SSM through learned partitions \cite{quadmamba}, predicted continuous offsets with bilinear interpolation \cite{damamba,defmamba}, content-dependent reordering \cite{objectness_scan,asm_unet,mambamatcher}, dynamic state aggregation \cite{pvmamba}, structure-aware state fusion \cite{spatialmamba}, non-causal SSD \cite{vssd}, or Mamba--Transformer hybridization \cite{mambavision}. GraphScan instead performs one-step semantic message passing on a local graph over visual tokens before the SSM: input-adaptive like dynamic scans, spatially bounded like local scans, and semantic like graph aggregation, while preserving token count and linear scaling.

\section{Method}
\label{sec:method}

GraphScan is designed around a simple principle: before a one-dimensional selective SSM performs global sequence mixing, each visual token should first be enriched by the local semantic neighborhood from which it arises. In this section, we first review the Vision Mamba selective-scan interface, then reinterpret adaptive scanning as pre-SSM token routing, introduce the GraphScan operator, and finally describe its integration into the GraphScan-Mamba backbone.

\subsection{Vision Selective Scan Preliminaries}
\label{sec:prelim}

Let $X\in\mathbb{R}^{H\times W\times D}$ denote an intermediate visual feature map with $D$ channels. We flatten it into a token sequence $\mathbf{X}=[\mathbf{x}_1,\ldots,\mathbf{x}_L]\in\mathbb{R}^{L\times D}$ with $L=HW$. Each token $i$ has a discrete lattice coordinate $\mathbf{g}_i\in\{1,\ldots,H\}\times\{1,\ldots,W\}$ and a normalized coordinate $\mathbf{p}_i\in[-1,1]^2$. The downstream SSM consumes the flattened sequence in raster order unless otherwise specified. Batch dimensions are omitted for clarity.

\noindent\textbf{Selective SSM interface.}
A Mamba-style selective SSM maps an input sequence $\mathbf{U}=[\mathbf{u}_1,\ldots,\mathbf{u}_L]$ to an output sequence through an input-dependent recurrence \cite{mamba}. Let $N_s$ denote the SSM state dimension. For the Mamba-1/S6 interface used by most Vision Mamba backbones, each channel maintains an $N_s$-dimensional recurrent state. At position $t$, the static transition parameter $\mathbf{A}\in\mathbb{R}^{D\times N_s}$ is combined with input-dependent quantities generated from $\mathbf{u}_t$: a final positive step size $\boldsymbol{\Delta}_t\in\mathbb{R}^{D}$ and selective vectors $\mathbf{b}_t,\mathbf{c}_t\in\mathbb{R}^{N_s}$. After discretization, this yields effective tensors $\bar{\mathbf{A}}_t,\bar{\mathbf{B}}_t\in\mathbb{R}^{D\times N_s}$, where $\bar{\mathbf{A}}_t$ controls state retention and $\bar{\mathbf{B}}_t$ controls the input write into the state. The recurrent state is $\mathbf{H}_t\in\mathbb{R}^{D\times N_s}$. We use two broadcast operators. For $\mathbf{u}_t\in\mathbb{R}^{D}$, let $\mathbf{u}_t^{\uparrow_s}\in\mathbb{R}^{D\times N_s}$ denote broadcasting over the state dimension, i.e., $[\mathbf{u}_t^{\uparrow_s}]_{d,n}=\mathbf{u}_{t,d}$. For $\mathbf{c}_t\in\mathbb{R}^{N_s}$, let $\mathbf{c}_t^{\uparrow_D}\in\mathbb{R}^{D\times N_s}$ denote broadcasting over the channel dimension, i.e., $[\mathbf{c}_t^{\uparrow_D}]_{d,n}=\mathbf{c}_{t,n}$. The core update is
\begin{equation}
\mathbf{H}_t
=
\bar{\mathbf{A}}_t \odot \mathbf{H}_{t-1}
+
\bar{\mathbf{B}}_t \odot \mathbf{u}_t^{\uparrow_s},
\qquad
\mathbf{y}_t
=
\left\langle \mathbf{H}_t,\mathbf{c}_t^{\uparrow_D}\right\rangle_{N_s},
\label{eq:ssm_recurrence}
\end{equation}
where $\langle\cdot,\cdot\rangle_{N_s}$ denotes contraction over the SSM state dimension, producing $\mathbf{y}_t\in\mathbb{R}^{D}$. Across the full sequence, the post-discretization tensors have shape $\bar{\mathbf{A}},\bar{\mathbf{B}}\in\mathbb{R}^{L\times D\times N_s}$, and the output sequence satisfies $\mathbf{Y}\in\mathbb{R}^{L\times D}$.

The key property for our work is selectivity: $\boldsymbol{\Delta}_t$, $\mathbf{b}_t$, and $\mathbf{c}_t$ are generated from the current token $\mathbf{u}_t$ \cite{mamba}. Therefore, modifying the token sequence before the SSM changes both the values written into the recurrent state and the selective parameters that govern propagation and readout. Mamba-2 and later SSD-style models provide alternative formulations and more efficient algorithms for related state-space sequence transformations \cite{mamba2}; our method targets the input sequence presented to the SSM and is therefore orthogonal to these recurrence-level developments.

\noindent\textbf{Parallel scan computation.}
Although Eq.~\eqref{eq:ssm_recurrence} is written recurrently, selective scan is computed efficiently using an associative prefix-scan operator \cite{mamba}. Define the write term $\mathbf{U}_t=\bar{\mathbf{B}}_t\odot\mathbf{u}_t^{\uparrow_s}$ and the affine update pair $\mathbf{P}_t=(\bar{\mathbf{A}}_t,\mathbf{U}_t)$. Composing two updates gives $(\mathbf{A}_2,\mathbf{U}_2)\circ(\mathbf{A}_1,\mathbf{U}_1) = (\mathbf{A}_2\odot\mathbf{A}_1,\;\mathbf{A}_2\odot\mathbf{U}_1+\mathbf{U}_2)$, which is associative. Thus all hidden states can be obtained by a parallel prefix scan over $\{\mathbf{P}_t\}_{t=1}^{L}$, preserving linear scaling in sequence length while avoiding materializing a dense $L\times L$ mixing matrix.

\noindent\textbf{Unrolled view.}
Unrolling Eq.~\eqref{eq:ssm_recurrence} from a zero initial state yields $\mathbf{H}_t = \sum_{i=1}^{t} \mathbf{G}_{t,i}\odot \bar{\mathbf{B}}_i\odot \mathbf{u}_i^{\uparrow_s}$ with $\mathbf{G}_{t,i} = \bigodot_{j=i+1}^{t}\bar{\mathbf{A}}_j$ the selective survival factor from sequence position $i$ to $t$. With a nonzero initial state, an additional term $\mathbf{G}_{t,0}\odot\mathbf{H}_0$ appears, where $\mathbf{G}_{t,0}=\bigodot_{j=1}^{t}\bar{\mathbf{A}}_j$; for standard visual prefill over a feature map, we use the zero-state form for clarity. For visual features, this one-dimensional path is only an indirect representation of the original two-dimensional structure. A pair of patches can be adjacent in the image but distant in raster order, making the scan/routing operator a critical interface between visual geometry and SSM computation.

\subsection{From Geometric Scanning to Semantic Routing}
\label{sec:geo_to_graph}

Existing Vision Mamba backbones improve this interface by changing how the image is serialized or resampled before selective scan. Fixed scans choose a deterministic path through the lattice; local scans restrict processing to windows; adaptive coordinate-based scans predict new sampling locations before feeding features into the SSM \cite{vim,vmamba,localmamba,damamba,defmamba}. Our goal is not to modify the SSM recurrence itself, but to improve the visual token sequence that enters it.

\noindent\textbf{Geometric adaptive scan as local interpolation.}
Coordinate-based adaptive scans predict an offset $\Delta\mathbf{p}_i$ for each reference position and sample from $\mathbf{p}'_i=\mathbf{p}_i+\Delta\mathbf{p}_i$. The output token is obtained through bilinear interpolation, $\mathbf{x}'_i = \sum_{j\in\mathcal{N}_4(\mathbf{p}'_i)} w^{\mathrm{bilin}}_{ij}\,\mathbf{x}_j$ with $w^{\mathrm{bilin}}_{ij}\ge 0$ and $\sum_j w^{\mathrm{bilin}}_{ij}=1$, where $\mathcal{N}_4(\mathbf{p}'_i)$ denotes the four lattice points surrounding the sampled coordinate. This is an effective form of input-dependent local rerouting, but the neighborhood and weights are determined by geometry: the predicted coordinate chooses the four corners and the interpolation rule fixes their weights.

\noindent\textbf{Our view.}
We instead ask whether adaptive scanning can be made semantic. Rather than predicting where to sample in continuous coordinate space, GraphScan asks which nearby patches should communicate with the current token. It preserves the same interface as prior pre-SSM rerouting operators: the output remains a length-$L$ sequence in the original raster slots. However, the content assigned to each slot is produced by learned message passing over a local graph, with weights determined by inter-patch affinities.

\subsection{GraphScan: Local Semantic Routing}
\label{sec:graphscan}

\begin{figure}[t]
    \centering
    \begin{minipage}[c]{0.72\textwidth}
        \centering
        \includegraphics[width=\linewidth]{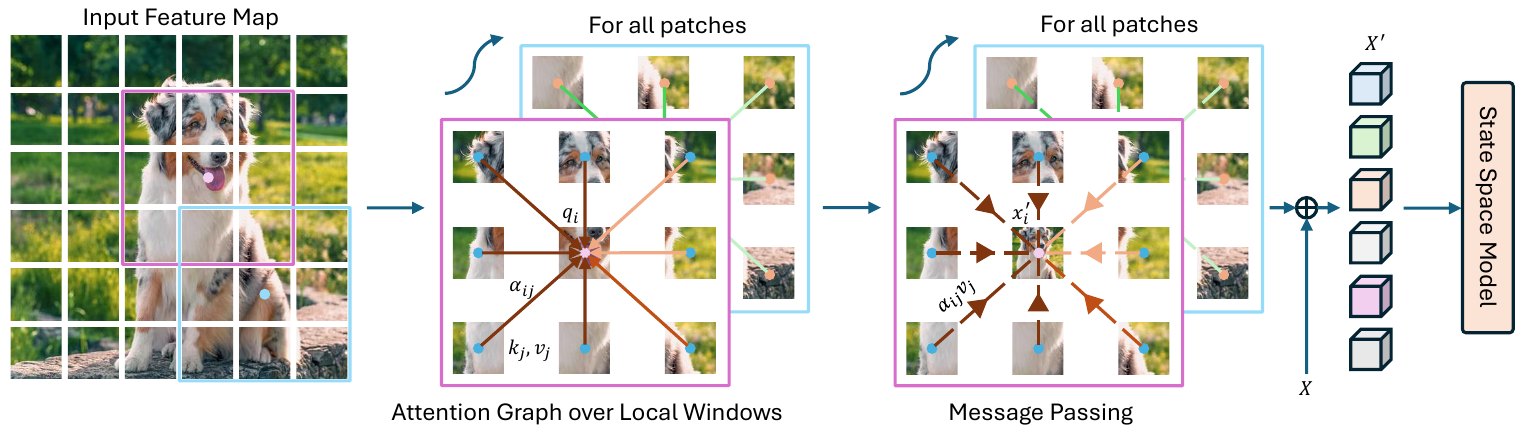}
    \end{minipage}
    \hfill
    \begin{minipage}[c]{0.26\textwidth}
       \caption{\footnotesize 
\textbf{GraphScan.}
Each token attends to a bounded local neighborhood, aggregates semantically weighted neighboring features, and produces a routed token \(x'_i\). The routed feature map \(X'\) is then passed to S6.
}
        \label{fig:graphscan}
    \end{minipage}
\end{figure}

GraphScan constructs, for each token, a bounded local graph on the image lattice and performs one step of feature-conditioned message passing \cite{gilmer2017mpnn}. It is deliberately lightweight: the neighborhood size is fixed by a small radius, the attention is local rather than global, and the output token count is unchanged. Fig.~\ref{fig:graphscan} illustrates the local affinity construction and message-passing update used to produce the routed token sequence.

\noindent\textbf{Local candidate set.}
For token $i$, we define a spatial candidate set $\mathcal{S}_r(i) = \{j : \|\mathbf{g}_j-\mathbf{g}_i\|_{\infty}\le r\}$ of size $S=|\mathcal{S}_r(i)|=(2r+1)^2$, where $r$ is the graph radius. In implementation, boundary tokens use replicate padding through coordinate clamping, so every token has the same $S$ window slots. This avoids introducing artificial zero tokens into the softmax and keeps the operator shape-static.

\noindent\textbf{Feature-conditioned affinities.}
For each token, GraphScan computes query, key, and value projections $\mathbf{q}_i=\mathbf{x}_iW_q$, $\mathbf{k}_j=\mathbf{x}_jW_k$, and $\mathbf{v}_j=\mathbf{x}_jW_v$, where $\mathbf{q}_i,\mathbf{k}_j\in\mathbb{R}^{d}$ and $\mathbf{v}_j\in\mathbb{R}^{d_v}$. For $j\in\mathcal{S}_r(i)$, the affinity score is
\begin{equation}
 s_{ij}
=
\frac{\mathbf{q}_i\mathbf{k}_j^{\top}}{\sqrt{d}}
+
 b_{\mathrm{rel}}(\mathbf{g}_j-\mathbf{g}_i),
\qquad
\alpha_{ij}
=
\frac{\exp(s_{ij})}{\sum_{\ell\in\mathcal{S}_r(i)}\exp(s_{i\ell})}.
\label{eq:graphscan_affinity}
\end{equation}
The learnable relative bias $b_{\mathrm{rel}}$ is defined over discrete offsets inside the local window. Unless otherwise stated, we use a single affinity head; a multi-head variant follows the standard head-splitting formulation and applies Eq.~\eqref{eq:graphscan_affinity} independently per head, with a shared relative-position bias $b_{\mathrm{rel}}$ across heads to keep parameter shapes invariant under head count.

\noindent\textbf{Message-passing update.}
The routed token is produced by aggregating value features from the learned semantic neighborhood:
\begin{equation}
\mathbf{x}'_i
=
\mathbf{x}_i
+
\left(
\sum_{j\in\mathcal{S}_r(i)}
\alpha_{ij}\,\mathbf{v}_j
\right)W_o,
\label{eq:graphscan_update}
\end{equation}
where $W_o\in\mathbb{R}^{d_v\times D}$ projects the aggregated message back to the model dimension. Eq.~\eqref{eq:graphscan_update} preserves the original token slot $i$, but replaces an isolated patch representation with a semantic mixture of nearby visual evidence. 
The local window specifies only where this evidence may come from; it is not a convolutional kernel. 
While convolution uses content-independent weights tied to relative offsets, GraphScan recomputes token-specific weights from inter-patch affinities. 
The operation is therefore related at the primitive level to local neighborhood attention and graph message passing over visual tokens \cite{nat,nonlocal,gat,gilmer2017mpnn,vig,pvg}, but its architectural role is distinct. 
GraphScan is not a standalone attention block or Vision-GNN backbone: it is a scan-time routing layer placed immediately before the selective SSM.
The routed tokens consequently affect both the values written into the recurrent state and the input-dependent parameters that govern state propagation and readout.
GraphScan therefore turns local semantic affinity into a preconditioning step for global state-space mixing (Sec.~\ref{sec:preconditioning}); a more detailed comparison to neighborhood attention and Vision-GNN primitives is given in Appendix~\ref{app:relation_local_attention_gnn}.

\noindent\textbf{Graph-induced scan field.}
For analysis and visualization, GraphScan induces an effective routed coordinate by taking the expected source position under the learned affinities, $\widehat{\mathbf{p}}_i = \sum_{j\in\mathcal{S}_r(i)} \alpha_{ij}\mathbf{p}_j$, with displacement $\Delta\widehat{\mathbf{p}}_i = \widehat{\mathbf{p}}_i-\mathbf{p}_i$. Since the weights $\alpha_{ij}$ are nonnegative and sum to one, $\widehat{\mathbf{p}}_i$ remains inside the image coordinate domain whenever all $\mathbf{p}_j\in[-1,1]^2$. This yields a displacement-field visualization analogous to offset-based scans (Fig.~\ref{fig:stage4_routing} in App.~\ref{app:visualizations}), but the displacement is induced by semantic routing rather than by direct coordinate regression.

\noindent\textbf{Complexity.}
With fixed radius $r$, GraphScan has constant graph degree $S=(2r+1)^2$. The local affinity and aggregation cost scales as $\mathcal{O}(LS(d+d_v))$ up to projection costs, and is therefore linear in the number of image tokens. The routed sequence $\mathbf{X}'=[\mathbf{x}'_1,\ldots,\mathbf{x}'_L]$ has the same length as $\mathbf{X}$ and can be passed directly to the selective SSM without changing the scan kernel.

\subsection{GraphScan as Selective-Scan Preconditioning}
\label{sec:preconditioning}

GraphScan can be viewed as a sparse, input-adaptive preconditioner for selective scan. 
Let $P(\mathbf{X})\in\mathbb{R}^{L\times L}$ denote the sparse row-stochastic routing matrix induced by Eq.~\eqref{eq:graphscan_affinity}, with $P_{ij}(\mathbf{X})=\alpha_{ij}$ for $j\in\mathcal{S}_r(i)$ and zero otherwise. 
Writing $M=W_vW_o$, Eq.~\eqref{eq:graphscan_update} becomes
\begin{equation}
\mathbf{X}'
=
\mathbf{X}
+
P(\mathbf{X})\mathbf{X}M,
\qquad
\mathbf{Y}
=
\mathrm{SSM}(\mathbf{X}').
\label{eq:graphscan_preconditioner}
\end{equation}
Thus, GraphScan changes the sequence presented to the SSM while leaving the selective-scan recurrence and implementation unchanged. 
The dependence of $P$ on $\mathbf{X}$ flows through the $W_q$ and $W_k$ projections inside the affinity computation in Eq.~\eqref{eq:graphscan_affinity}.

\noindent\textbf{Parameter-modulated selective scan.}
The effect of Eq.~\eqref{eq:graphscan_preconditioner} is not limited to changing token values. 
Since selective SSMs generate $\boldsymbol{\Delta}_t$, $\bar{\mathbf{B}}_t$, and $\mathbf{c}_t$ from the current token, replacing $\mathbf{x}_t$ with $\mathbf{x}'_t$ also changes the input-dependent parameters of the recurrence. 
The routed recurrence is therefore
\begin{equation}
\begin{aligned}
\mathbf{H}'_t
&=
\bar{\mathbf{A}}'_t\odot \mathbf{H}'_{t-1}
+
\bar{\mathbf{B}}'_t\odot(\mathbf{x}'_t)^{\uparrow_s},\\
\mathbf{y}'_t
&=
\left\langle
\mathbf{H}'_t,
(\mathbf{c}'_t)^{\uparrow_D}
\right\rangle_{N_s},
\end{aligned}
\label{eq:routed_ssm}
\end{equation}
where $\bar{\mathbf{A}}'_t=\bar{\mathbf{A}}_t(\mathbf{x}'_t)$, $\bar{\mathbf{B}}'_t=\bar{\mathbf{B}}_t(\mathbf{x}'_t)$, and $\mathbf{c}'_t=\mathbf{c}_t(\mathbf{x}'_t)$. 
This is the central reason for placing GraphScan immediately before the SSM: local semantic routing affects both the values written into the recurrent state and the token-dependent parameters that govern propagation and readout.

\noindent\textbf{Exact routed-vs-base decomposition.}
Let 
\[
\mathbf{m}_t=\mathbf{x}'_t-\mathbf{x}_t
=
\sum_{j\in\mathcal{S}_r(t)}\alpha_{tj}\mathbf{x}_jM
\]
denote the GraphScan message added to token $t$. 
Define the exact parameter differences 
$\delta\bar{\mathbf{A}}_t=\bar{\mathbf{A}}'_t-\bar{\mathbf{A}}_t$, 
$\delta\bar{\mathbf{B}}_t=\bar{\mathbf{B}}'_t-\bar{\mathbf{B}}_t$, and 
$\delta\mathbf{c}_t=\mathbf{c}'_t-\mathbf{c}_t$, where unprimed quantities are generated from the base token $\mathbf{x}_t$. 
Comparing the routed recurrence in Eq.~\eqref{eq:routed_ssm} with the base recurrence in Eq.~\eqref{eq:ssm_recurrence} yields the exact decomposition
\begin{equation}
\delta\mathbf{H}_t
=
\sum_{i=1}^{t}
\mathbf{G}'_{t,i}\odot
\left[
\bar{\mathbf{B}}'_i\odot\mathbf{m}_i^{\uparrow_s}
+
\delta\bar{\mathbf{B}}_i\odot\mathbf{x}_i^{\uparrow_s}
+
\delta\bar{\mathbf{A}}_i\odot\mathbf{H}_{i-1}
\right],
\label{eq:exact_state_decomp}
\end{equation}
where $\delta\mathbf{H}_t=\mathbf{H}'_t-\mathbf{H}_t$ and 
$\mathbf{G}'_{t,i}=\bigodot_{j=i+1}^{t}\bar{\mathbf{A}}'_j$ is the routed survival factor. 
The corresponding output difference is
\begin{equation}
\delta\mathbf{y}_t
=
\left\langle
\delta\mathbf{H}_t,
(\mathbf{c}'_t)^{\uparrow_D}
\right\rangle_{N_s}
+
\left\langle
\mathbf{H}_t,
\delta\mathbf{c}_t^{\uparrow_D}
\right\rangle_{N_s}.
\label{eq:exact_output_decomp}
\end{equation}
Eq.~\eqref{eq:exact_state_decomp} separates three hidden-state effects of GraphScan: direct routed value injection through $\bar{\mathbf{B}}'_i\odot\mathbf{m}_i^{\uparrow_s}$, write modulation through $\delta\bar{\mathbf{B}}_i$, and transition modulation through $\delta\bar{\mathbf{A}}_i$. 
Eq.~\eqref{eq:exact_output_decomp} further adds readout modulation through $\delta\mathbf{c}_t$. 
Thus, GraphScan does not merely add a local message before the SSM; it changes how the selective scan writes, retains, and reads visual information. 
The full derivation is given in Appendix~\ref{app:routed_decomposition}.

\noindent\textbf{Local-global kernel view.}
The direct routed value path in Eq.~\eqref{eq:exact_state_decomp} exposes the local-global structure induced by GraphScan. 
Define the routed SSM input-to-output operator
\begin{equation}
\mathcal{K}'_{t,i}(\mathbf{z})
=
\left\langle
\mathbf{G}'_{t,i}\odot\bar{\mathbf{B}}'_i\odot\mathbf{z}^{\uparrow_s},
(\mathbf{c}'_t)^{\uparrow_D}
\right\rangle_{N_s}.
\label{eq:routed_kernel_operator}
\end{equation}
Using $\mathbf{m}_i=\sum_{j\in\mathcal{S}_r(i)}\alpha_{ij}\mathbf{x}_jM$, the direct routed value contribution becomes
\begin{equation}
\delta\mathbf{y}^{\mathrm{route}}_t
=
\sum_{j=1}^{L}
\sum_{\substack{i\le t:\\ j\in\mathcal{S}_r(i)}}
\alpha_{ij}\,
\mathcal{K}'_{t,i}(\mathbf{x}_jM).
\label{eq:exact_local_global_kernel}
\end{equation}
A source patch $\mathbf{x}_j$ is therefore weighted twice: first by a local semantic affinity $\alpha_{ij}$ on the image lattice, and then by the global selective propagation encoded in $\mathcal{K}'_{t,i}$. 
GraphScan consequently creates a local semantic shortcut before the one-dimensional recurrence, allowing spatially nearby or semantically related patches to influence the selective scan without relying solely on their raster-order distance. 
The corresponding derivation is given in Appendix~\ref{app:local_global_kernel}.

\noindent\textbf{Further structural consequences.}
The preconditioner view yields three additional consequences detailed in App.~\ref{app:graphscan_analysis}: a scan-distance attenuation bound that governs how a source patch reaches output $t$ through a routed slot $i$ in terms of $t-i$ rather than raster distance (App.~\ref{app:scan_distance}); a modified finite-horizon input-to-state map with cross-token directions induced by the sparse routing matrix $P(\mathbf{X})$ (App.~\ref{app:reachability}); and baseline containment---setting $W_o=0$ recovers $\mathbf{X}'=\mathbf{X}$ (App.~\ref{app:containment}). GraphScan therefore augments the backbone's representational family while preserving the original Vision Mamba model as a special case.

\noindent\textbf{Orthogonality to SSM cores.}
GraphScan is an input-side operator that leaves the discretization rule, selective scan recurrence, SSD formulation, and hidden-state update untouched, so it is complementary to recurrence-side improvements such as SSD, non-causal SSD, and newer Mamba variants \cite{mamba2,vssd,mamba3}. We instantiate it on the Mamba-1/S6 Vision Mamba interface to isolate the effect of scanning and enable direct comparison with existing scan mechanisms.

\subsection{GraphScan-Mamba Architecture}
\label{sec:architecture}

\begin{figure}[t]
    \centering
    \includegraphics[width=0.9\textwidth]{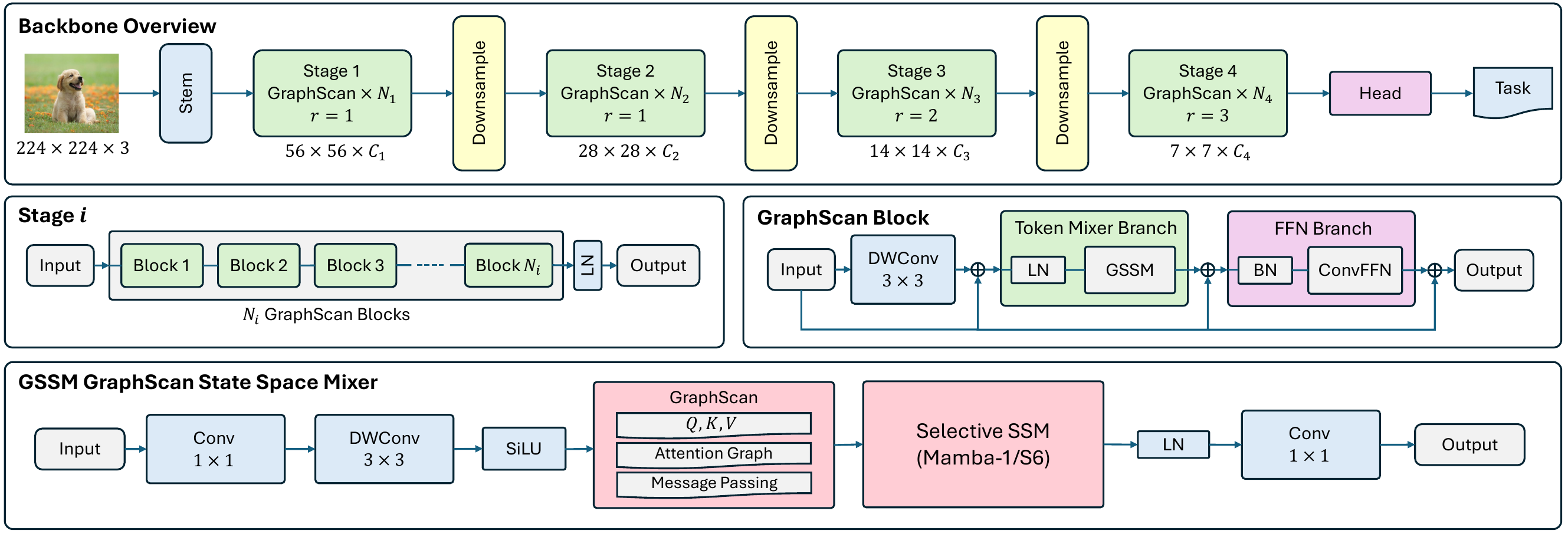}
    \caption{\textbf{GraphScan-Mamba architecture.} \emph{Top:} four-stage hierarchical backbone with an overlapping convolutional stem to $1/4$ resolution and stage-wise downsampling, with stage-wise graph radii $r=(1,1,2,3)$ yielding $3{\times}3$, $3{\times}3$, $5{\times}5$, and $7{\times}7$ local neighborhoods. \emph{Middle:} each of $N_i$ blocks combines a depthwise positional branch, a GraphScan-SSM token-mixing branch, and a ConvFFN branch. \emph{Bottom:} inside GSSM, features pass through $1{\times}1$ projection, $3{\times}3$ depthwise conv, SiLU, GraphScan, selective SSM, normalization, and output projection---a fixed-local $\to$ learned-local $\to$ global progression.}
    \label{fig:architecture}
\end{figure}
\vspace{-2mm}

We integrate GraphScan into a hierarchical Vision Mamba backbone, \textbf{GraphScan-Mamba}, following the standard four-stage design used by modern vision backbones \cite{vmamba,localmamba,msvmamba,spatialmamba,damamba}: an overlapping convolutional stem produces a $1/4$-resolution feature map, and subsequent stages progressively downsample spatial resolution while increasing channel dimension. This architectural template is deliberately conventional; our goal is to isolate the effect of replacing geometric pre-SSM routing with graph-induced semantic routing. Fig.~\ref{fig:architecture} summarizes the full backbone, the repeated GraphScan block, and the GSSM token mixer where GraphScan is inserted before the selective SSM.

\noindent\textbf{Block structure.}
Each block has three residual components applied in sequence: $X \leftarrow X + \mathrm{DWConv}_{3\times3}(X)$, then $X \leftarrow X + \mathrm{GSSM}(\mathrm{LN}(X))$, then $X \leftarrow X + \mathrm{ConvFFN}(\mathrm{BN}(X))$ (omitting drop path and layer-scale factors). The first is a residual depthwise convolution that injects local positional bias \cite{cpvt,damamba}, fused at implementation time into a single convolution with an identity-offset kernel. $\mathrm{ConvFFN}$ stacks a $1{\times}1$ pointwise convolution, a $3{\times}3$ depthwise convolution, and a final $1{\times}1$ pointwise convolution \cite{pvtv2,damamba}.

\noindent\textbf{GraphScan-SSM token mixer.}
GraphScan is inserted between the input stage and the selective SSM: $\widetilde{X} = \mathrm{SiLU}(\mathrm{DWConv}_{3\times3}(\mathrm{Proj}_{\mathrm{in}}(X)))$, $X^{\mathrm{gs}} = \mathrm{GraphScan}(\widetilde{X})$, and $\mathrm{GSSM}(X) = \mathrm{Proj}_{\mathrm{out}}(\mathrm{LN}(\mathrm{SSM}(X^{\mathrm{gs}})))$. Each token mixer therefore follows a \emph{fixed-local} (depthwise convolution) $\to$ \emph{learned-local} (GraphScan) $\to$ \emph{global} (selective SSM) progression.

\noindent\textbf{Stage-wise graph radius.}
The graph radius $r$ grows with stage depth: smaller windows at high resolution, where local mixing suffices, and larger windows in low-resolution late stages, where the receptive field needs to become effectively global. At the final stage, the GraphScan neighborhood spans the entire feature map while remaining linear in the spatial budget. Exact radii, channel widths, depths, and training configurations are reported in Sec.~\ref{sec:exp}.

\section{Experiments}
\label{sec:exp}

We evaluate GraphScan-Mamba on three standard recognition benchmarks: ImageNet-1K classification \cite{imagenet}, COCO object detection and instance segmentation \cite{coco} with Mask R-CNN \cite{maskrcnn}, and ADE20K semantic segmentation \cite{ade20k} with UperNet \cite{upernet}. Across all three tasks, GraphScan-Mamba achieves state-of-the-art accuracy among Vision SSM backbones at comparable parameter and FLOP budgets. We also ablate pre-SSM placement, graph radius, relative positional bias, and head count; per-block attention and effective-routing visualizations of the trained backbone are reported in App.~\ref{app:visualizations}.

\noindent\textbf{Architecture variants.}
We instantiate three model sizes (Tiny, Small, Base), following the conventional channel-and-depth scaling axes used by hierarchical Vision Mamba backbones \cite{vmamba}. The macro topology, MLP ratios, and graph radii are shared across variants; only channel widths and stage depths vary. Radii follow $r=(1,1,2,3)$, yielding $3{\times}3$, $3{\times}3$, $5{\times}5$, and $7{\times}7$ neighborhoods at stages 1--4; the stage-4 neighborhood spans the full $7{\times}7$ feature map. Per-variant channel widths, depths, MLP ratios, parameters, and FLOPs are reported in App.~\ref{app:variants}.

\subsection{Image Classification on ImageNet-1K}
\label{sec:exp_in1k}
\vspace{-2mm}

\noindent
\begin{minipage}[t]{0.50\textwidth}
\vspace{0pt}
\noindent\textbf{Setup.}
We follow the standard DeiT-style training recipe \cite{deit} adopted across recent Vision Mamba works \cite{vmamba}. All variants are trained for 300 epochs at $224^2$ resolution using AdamW with a cosine learning-rate schedule, 20-epoch linear warmup, base learning rate $1{\times}10^{-3}$, weight decay $5{\times}10^{-2}$, and batch size 1024. Augmentation and regularization include RandAugment, Repeated Augmentation, Mixup, CutMix, Random Erasing, label smoothing, and stochastic depth, with the maximum stochastic-depth rate linearly interpolated to $\{0.2, 0.3, 0.5\}$ for T/S/B. Center cropping is applied at evaluation. FLOPs and parameters are reported at $224^2$.

\vspace{0.6em}
\noindent\textbf{Results.}
As shown in Table~\ref{tab:in1k}, GraphScan-Mamba consistently outperforms ConvNet, Transformer, and Vision SSM backbones at every parameter scale. GraphScan-Mamba-B reaches $86.5\%$ top-1 accuracy, $+1.2$ over the strongest Vision SSM at this scale (Spatial-Mamba-B at $85.3\%$) and $+2.6$ over VMamba-B at $83.9\%$. At the Small scale, GraphScan-Mamba-S obtains $85.7\%$, exceeding DAMamba-S ($84.8\%$) by $+0.9$ and VMamba-S ($83.6\%$) by $+2.1$. At Tiny, GraphScan-Mamba-T attains $84.4\%$, $+0.6$ above the strongest Vision SSM at this scale (DAMamba-T at $83.8\%$) and $+1.9$ above VMamba-T at $82.5\%$. The lead is preserved against the strongest ConvNets and Transformers at matched parameter budgets, including ConvNeXt-B ($83.8\%$), Swin-B ($83.5\%$), and NAT-B ($84.3\%$). These results demonstrate that injecting a local semantic routing layer before the selective scan is an effective lever for surpassing strong ConvNet and Transformer backbones while preserving the linear scaling of state-space models.
\end{minipage}%
\hfill
\begin{minipage}[t]{0.48\textwidth}
\vspace{0pt}
\centering
\captionof{table}{\textbf{Image classification on ImageNet-1K.} All entries use $224^2$ input. ``Type'': ConvNet (C), Transformer (T), State-Space Model (S), or Graph Neural Network (G). Best result per parameter regime is in \textbf{bold}.}
\label{tab:in1k}
\resizebox{\linewidth}{!}{%
\renewcommand{\arraystretch}{0.85}
\setlength{\tabcolsep}{3pt}
\begin{tabular}{lcccc}
\toprule
Method & Type & Params & FLOPs & Top-1 (\%) \\
\midrule
ConvNeXt-T \cite{convnext}        & C & 29M & 4.5G  & 82.1 \\
InceptionNeXt-T \cite{inceptionnext} & C & 28M & 4.2G  & 82.3 \\
Swin-T \cite{swin}                & T & 29M & 4.5G  & 81.3 \\
NAT-T \cite{nat}                  & T & 28M & 4.3G  & 83.2 \\
BiFormer-S \cite{biformer}        & T & 26M & 4.5G  & 83.8 \\
PyramidViG-S \cite{vig} & G & 27M & 4.6G & 82.1 \\
VCMamba-M \cite{vcmamba} & S & 21M & 2.3G & 81.5 \\
VCMamba-B \cite{vcmamba} & S & 32M & 4.0G & 82.6 \\
Vim-S \cite{vim}                  & S & 26M & 5.1G  & 80.5 \\
PlainMamba-L2 \cite{plainmamba}   & S & 25M & 8.1G  & 81.6 \\
VMamba-T \cite{vmamba}            & S & 31M & 4.9G  & 82.5 \\
LocalVMamba-T \cite{localmamba}   & S & 26M & 5.7G  & 82.7 \\
MSVMamba-T \cite{msvmamba}        & S & 33M & 4.6G  & 82.8 \\
MambaVision-T \cite{mambavision}  & S & 32M & 4.4G  & 82.3 \\
GroupMamba-T \cite{groupmamba}    & S & 23M & 4.5G  & 83.3 \\
GroupMamba-S \cite{groupmamba}    & S & 34M & 7.0G  & 83.9 \\
Spatial-Mamba-T \cite{spatialmamba} & S & 27M & 4.5G & 83.5 \\
VSSD-T \cite{vssd}                & S & 24M & 4.5G  & 83.7 \\
DAMamba-T \cite{damamba}          & S & 26M & 4.8G  & 83.8 \\
\rowcolor{gray!12}\textbf{GraphScan-Mamba-T (ours)} & S & 28M & 5.2G & \textbf{84.4} \\
\midrule
ConvNeXt-S \cite{convnext}        & C & 50M &  8.7G & 83.1 \\
InceptionNeXt-S \cite{inceptionnext} & C & 49M &  8.4G & 83.5 \\
Swin-S \cite{swin}                & T & 50M &  8.7G & 83.0 \\
NAT-S \cite{nat}                  & T & 51M &  7.8G & 83.7 \\
BiFormer-B \cite{biformer}        & T & 57M &  9.8G & 84.3 \\
PyramidViG-M \cite{vig} & G & 52M & 8.9G & 83.1 \\
PlainMamba-L3 \cite{plainmamba}   & S & 50M & 14.4G & 82.3 \\
VMamba-S \cite{vmamba}            & S & 50M &  8.7G & 83.6 \\
LocalVMamba-S \cite{localmamba}   & S & 50M & 11.4G & 83.7 \\
MambaVision-S \cite{mambavision}  & S & 50M &  7.5G & 83.3 \\
VSSD-S \cite{vssd}                & S & 40M &  7.4G & 84.1 \\
Spatial-Mamba-S \cite{spatialmamba} & S & 43M &  7.1G & 84.6 \\
DAMamba-S \cite{damamba}          & S & 45M & 10.3G & 84.8 \\
\rowcolor{gray!12}\textbf{GraphScan-Mamba-S (ours)} & S & 49M & 11.1G & \textbf{85.7} \\
\midrule
ConvNeXt-B \cite{convnext}        & C & 89M & 15.4G & 83.8 \\
InceptionNeXt-B \cite{inceptionnext} & C & 87M & 14.9G & 84.0 \\
Swin-B \cite{swin}                & T & 88M & 15.4G & 83.5 \\
NAT-B \cite{nat}                  & T & 90M & 13.7G & 84.3 \\
PyramidViG-B \cite{vig} & G & 93M & 16.8G & 83.7 \\
GroupMamba-B \cite{groupmamba}    & S & 57M & 14.0G & 84.5 \\
VMamba-B \cite{vmamba}            & S & 89M & 15.4G & 83.9 \\
MambaVision-B \cite{mambavision}  & S & 98M & 15.0G & 84.2 \\
VSSD-B \cite{vssd}                & S & 89M & 16.1G & 84.7 \\
DAMamba-B \cite{damamba}          & S & 86M & 16.3G & 85.2 \\
Spatial-Mamba-B \cite{spatialmamba} & S & 96M & 15.8G & 85.3 \\
\rowcolor{gray!12}\textbf{GraphScan-Mamba-B (ours)} & S & 93M & 17.8G & \textbf{86.5} \\
\bottomrule
\end{tabular}}
\end{minipage}
\vspace{-1mm}

\subsection{Object Detection and Instance Segmentation on COCO}
\label{sec:exp_coco}

\noindent\textbf{Setup.}
We adopt Mask R-CNN \cite{maskrcnn} with FPN, training on COCO \texttt{train2017} and reporting on \texttt{val2017} \cite{coco}. We use the standard $1{\times}$ (12-epoch, single-scale) and $3{\times}$+MS (36-epoch, multi-scale) schedules with AdamW at initial learning rate $1{\times}10^{-4}$ and batch size 16. Backbones are initialized from ImageNet-1K pretraining; the rest of the detector is randomly initialized. All experiments are implemented in MMDetection \cite{mmdetection}. FLOPs are reported at $1280{\times}800$ input, and we report box AP ($\mathrm{AP^{b}}$), $\mathrm{AP^{b}_{50}}$, mask AP ($\mathrm{AP^{m}}$), and $\mathrm{AP^{m}_{50}}$ for both schedules.

\noindent\textbf{Results.}
Table~\ref{tab:coco} summarizes COCO performance. Under the $1{\times}$ schedule, GraphScan-Mamba-T/S/B achieve box mAP of $49.1/50.7/51.9$, outperforming VMamba-T/S/B by $1.8/2.0/2.7$, Swin-T/S/B by $6.4/5.9/5.0$, and ConvNeXt-T/S/B by $4.9/5.3/4.9$. Mask mAP follows the same trend at $43.9/45.3/46.0$, exceeding VMamba-T/S/B by $1.2/1.6/1.9$. Under the longer $3{\times}$+MS schedule, GraphScan-Mamba-T/S/B reach $51.0/52.0/52.5$ box mAP and $45.3/45.9/46.2$ mask mAP, again leading the strongest Vision SSM at each scale. The gains transfer to dense prediction without modifying the backbone, indicating that the local semantic routing learned during classification yields tokens that remain useful when the network is evaluated against detection-scale spatial structure.

\subsection{Semantic Segmentation on ADE20K}
\label{sec:exp_ade20k}

\noindent\textbf{Setup.}
We adopt UperNet \cite{upernet} on ADE20K \cite{ade20k}, fine-tuning ImageNet-pretrained backbones for 160K iterations at $512{\times}512$ crops with AdamW and batch size 16. All experiments are implemented in MMSegmentation \cite{mmsegmentation}. We report single-scale (SS) and multi-scale (MS) mIoU; FLOPs are reported at $512{\times}2048$ input.

\noindent\textbf{Results.}
Table~\ref{tab:ade20k} summarizes ADE20K accuracy. GraphScan-Mamba-T/S/B reach mIoU (SS) of $50.9/52.1/53.2$, surpassing VMamba-T/S/B by $2.9/1.5/2.2$ and the strongest Vision SSM baseline at each scale by $+0.6/+0.9/+1.3$. Multi-scale tests show analogous improvements ($51.8/52.9/53.6$). As discussed in Sec.~\ref{sec:preconditioning}, the local-global structure induced by GraphScan supplies semantically nearby content into the selective scan, which is especially valuable for scene-level segmentation, where two-dimensional adjacency frequently disagrees with raster order.

\begin{table}[t]
\centering
\begin{minipage}[t]{0.58\textwidth}
\centering
\caption{ \footnotesize \textbf{COCO object detection and instance segmentation} with Mask R-CNN, $1{\times}$ and $3{\times}$+MS schedules. ``\#P'' is the full-detector parameter count.}
\label{tab:coco}
\resizebox{\linewidth}{!}{%
\renewcommand{\arraystretch}{0.92}
\setlength{\tabcolsep}{2pt}
\begin{tabular}{lcccccccccc}
\toprule
& & \multicolumn{4}{c}{Mask R-CNN $1{\times}$} & & \multicolumn{4}{c}{Mask R-CNN $3{\times}$+MS} \\
\cmidrule{3-6}\cmidrule{8-11}
Backbone & \#P & $\mathrm{AP^{b}}$ & $\mathrm{AP^{b}_{50}}$ & $\mathrm{AP^{m}}$ & $\mathrm{AP^{m}_{50}}$ & & $\mathrm{AP^{b}}$ & $\mathrm{AP^{b}_{50}}$ & $\mathrm{AP^{m}}$ & $\mathrm{AP^{m}_{50}}$ \\
\midrule
Swin-T \cite{swin}                  & 48M & 42.7 & 65.2 & 39.3 & 62.2 & & 46.0 & 68.1 & 41.6 & 65.1 \\
ConvNeXt-T \cite{convnext}          & 48M & 44.2 & 66.6 & 40.1 & 63.3 & & 46.2 & 67.9 & 41.7 & 65.0 \\
NAT-T \cite{nat}                    & 48M & --   & --   & --   & --   & & 47.7 & 69.0 & 42.6 & 66.1 \\
BiFormer-S \cite{biformer}          & --  & 47.8 & 69.8 & 43.2 & 66.8 & & --   & --   & --   & --   \\
VMamba-T \cite{vmamba}              & 50M & 47.3 & 69.3 & 42.7 & 66.4 & & 48.8 & 70.4 & 43.7 & 67.4 \\
LocalVMamba-T \cite{localmamba}     & 45M & 46.7 & 68.7 & 42.2 & 65.7 & & 48.7 & 70.1 & 43.4 & 67.0 \\
MSVMamba-T \cite{msvmamba}          & 53M & 46.9 & 68.8 & 42.2 & 65.6 & & 48.3 & 69.5 & 43.2 & 66.8 \\
GroupMamba-T \cite{groupmamba}      & 40M & 47.6 & 69.8 & 42.9 & 66.5 & & --   & --   & --   & --   \\
VSSD-T \cite{vssd}                  & 44M & 46.9 & 69.4 & 42.6 & 66.4 & & 48.8 & 70.4 & 43.6 & 67.6 \\
Spatial-Mamba-T \cite{spatialmamba} & 46M & 47.6 & 69.6 & 42.9 & 66.5 & & 49.3 & 70.7 & 43.8 & 67.8 \\
DAMamba-T \cite{damamba}            & 45M & 48.5 & 70.3 & 43.4 & 67.2 & & 50.4 & 71.4 & 44.8 & 68.6 \\
\rowcolor{gray!12}\textbf{GraphScan-Mamba-T (ours)} & 47M & \textbf{49.1} & \textbf{70.9} & \textbf{43.9} & \textbf{67.7} & & \textbf{51.0} & \textbf{71.9} & \textbf{45.3} & \textbf{69.1} \\
\midrule
Swin-S \cite{swin}                  & 69M & 44.8 & 68.6 & 40.9 & 65.3 & & 48.2 & 69.8 & 43.2 & 67.0 \\
ConvNeXt-S \cite{convnext}          & 70M & 45.4 & 67.9 & 41.8 & 65.2 & & 47.9 & 70.0 & 42.9 & 66.9 \\
NAT-S \cite{nat}                    & 70M & --   & --   & --   & --   & & 48.4 & 69.8 & 43.2 & 66.9 \\
BiFormer-B \cite{biformer}          & --  & 48.6 & 70.5 & 43.7 & 67.6 & & --   & --   & --   & --   \\
VMamba-S \cite{vmamba}              & 70M & 48.7 & 70.0 & 43.7 & 67.3 & & 49.9 & 70.9 & 44.2 & 68.2 \\
LocalVMamba-S \cite{localmamba}     & 69M & 48.4 & 69.9 & 43.2 & 66.7 & & 49.9 & 70.5 & 44.1 & 67.8 \\
VSSD-S \cite{vssd}                  & 59M & 48.4 & 70.1 & 43.5 & 67.2 & & --   & --   & --   & --   \\
Spatial-Mamba-S \cite{spatialmamba} & 63M & 49.2 & 70.8 & 44.0 & 67.9 & & 50.5 & 71.5 & 44.6 & 68.7 \\
DAMamba-S \cite{damamba}            & 65M & 49.8 & 71.2 & 44.5 & 68.4 & & 51.2 & 72.1 & 45.1 & 69.2 \\
\rowcolor{gray!12}\textbf{GraphScan-Mamba-S (ours)} & 69M & \textbf{50.7} & \textbf{72.1} & \textbf{45.3} & \textbf{69.1} & & \textbf{52.0} & \textbf{72.8} & \textbf{45.9} & \textbf{69.9} \\
\midrule
Swin-B \cite{swin}                  & 107M & 46.9 & 69.2 & 42.3 & 66.0 & & 48.6 & 70.0 & 43.3 & 67.1 \\
ConvNeXt-B \cite{convnext}          & 107M & 47.0 & 69.4 & 42.7 & 66.3 & & 48.5 & 70.1 & 43.5 & 67.1 \\
VMamba-B \cite{vmamba}              & 108M & 49.2 & 71.4 & 44.1 & 68.3 & & --   & --   & --   & --   \\
Spatial-Mamba-B \cite{spatialmamba} & 115M & 50.4 & 71.8 & 45.1 & 69.1 & & --   & --   & --   & --   \\
DAMamba-B \cite{damamba}            & 105M & 50.6 & 71.9 & 44.9 & 68.9 & & 51.4 & 72.3 & 45.3 & 69.5 \\
\rowcolor{gray!12}\textbf{GraphScan-Mamba-B (ours)} & 112M & \textbf{51.9} & \textbf{73.0} & \textbf{46.0} & \textbf{69.9} & & \textbf{52.5} & \textbf{73.2} & \textbf{46.2} & \textbf{70.3} \\
\bottomrule
\end{tabular}}
\end{minipage}
\hfill
\begin{minipage}[t]{0.41\textwidth}
\centering
\caption{\footnotesize \textbf{Semantic segmentation on ADE20K} with UperNet, 160K iters. ``\#P'' and FLOPs at $512{\times}2048$.}
\label{tab:ade20k}
\resizebox{\linewidth}{!}{%
\renewcommand{\arraystretch}{0.92}
\setlength{\tabcolsep}{3pt}
\begin{tabular}{lcccc}
\toprule
Backbone & \#P & FLOPs & SS & MS \\
\midrule
Swin-T \cite{swin}                  & 60M & 945G  & 44.4 & 45.8 \\
ConvNeXt-T \cite{convnext}          & 60M & 939G  & 46.0 & 46.7 \\
NAT-T \cite{nat}                    & 58M & 934G  & 47.1 & 48.4 \\
BiFormer-S \cite{biformer}          & --  & --    & 49.8 & 50.8 \\
VMamba-T \cite{vmamba}              & 62M & 949G  & 48.0 & 48.8 \\
LocalVMamba-T \cite{localmamba}     & 57M & 970G  & 47.9 & 49.1 \\
MSVMamba-T \cite{msvmamba}          & 65M & 942G  & 47.6 & 48.5 \\
GroupMamba-T \cite{groupmamba}      & --  & --    & 48.6 & 49.2 \\
VSSD-T \cite{vssd}                  & 53M & 941G  & 47.9 & 48.7 \\
Spatial-Mamba-T \cite{spatialmamba} & 57M & 936G  & 48.6 & 49.4 \\
DAMamba-T \cite{damamba}            & 55M & 937G  & 50.3 & 51.2 \\
\rowcolor{gray!12}\textbf{GraphScan-Mamba-T (ours)} & 57M & 942G & \textbf{50.9} & \textbf{51.8} \\
\midrule
Swin-S \cite{swin}                  & 81M & 1039G & 47.6 & 49.5 \\
ConvNeXt-S \cite{convnext}          & 82M & 1027G & 48.7 & 49.6 \\
NAT-S \cite{nat}                    & 82M & 1010G & 48.0 & 49.5 \\
BiFormer-B \cite{biformer}          & --  & --    & 51.0 & 51.7 \\
VMamba-S \cite{vmamba}              & 82M & 1028G & 50.6 & 51.2 \\
LocalVMamba-S \cite{localmamba}     & 81M & 1095G & 50.0 & 51.0 \\
Spatial-Mamba-S \cite{spatialmamba} & 73M & 992G  & 50.6 & 51.4 \\
DAMamba-S \cite{damamba}            & 75M & 1050G & 51.2 & 52.0 \\
\rowcolor{gray!12}\textbf{GraphScan-Mamba-S (ours)} & 79M & 1065G & \textbf{52.1} & \textbf{52.9} \\
\midrule
Swin-B \cite{swin}                  & 121M & 1188G & 48.1 & 49.7 \\
ConvNeXt-B \cite{convnext}          & 122M & 1170G & 49.1 & 49.9 \\
NAT-B \cite{nat}                    & 123M & 1137G & 48.5 & 49.7 \\
VMamba-B \cite{vmamba}              & 122M & 1170G & 51.0 & 51.6 \\
Spatial-Mamba-B \cite{spatialmamba} & 127M & 1176G & 51.8 & 52.6 \\
DAMamba-B \cite{damamba}            & 117M & 1178G & 51.9 & 52.3 \\
\rowcolor{gray!12}\textbf{GraphScan-Mamba-B (ours)} & 124M & 1210G & \textbf{53.2} & \textbf{53.6} \\
\bottomrule
\end{tabular}}
\end{minipage}
\end{table}

\subsection{Ablation Studies}
\label{sec:exp_ablation}

We ablate four design choices on GraphScan-Mamba-T with all training settings inherited from Sec.~\ref{sec:exp_in1k}: pre- vs.\ post-SSM placement of the routing operator, graph radius schedule, the relative-position bias $b_{\mathrm{rel}}$, and the number of affinity heads. Each ablation modifies a single factor while holding the rest at their default; the default configuration is marked with $\star$.
\vspace{-2mm}

\begin{table}[h]
\centering
\footnotesize
\setlength{\tabcolsep}{4pt}
\caption{\footnotesize \textbf{Ablation studies} on GraphScan-Mamba-T (training settings identical to Sec.~\ref{sec:exp_in1k}). Default configuration marked with~$\star$. Top-1 (\%) reported.}
\label{tab:ablations}
\begin{minipage}[t]{0.24\textwidth}
\centering
(a) \textbf{Routing placement} \\[2pt]
\begin{tabular}{lc}
\toprule
Placement & Top-1 \\
\midrule
None (base SSM)            & 82.4 \\
Post-SSM                   & 83.7 \\
\rowcolor{gray!12}Pre-SSM~$\star$            & \textbf{84.4} \\
\bottomrule
\end{tabular}
\end{minipage}
\hfill
\begin{minipage}[t]{0.24\textwidth}
\centering
(b) \textbf{Radius schedule} \\[2pt]
\begin{tabular}{lc}
\toprule
$(r_1,r_2,r_3,r_4)$ & Top-1 \\
\midrule
$(1,1,1,1)$ & 83.5 \\
\rowcolor{gray!12}$(1,1,2,3)~\star$ & 84.4 \\
$(2,2,3,3)$ & \textbf{84.7} \\
\bottomrule
\end{tabular}
\end{minipage}
\hfill
\begin{minipage}[t]{0.22\textwidth}
\centering
(c) \textbf{Rel.\ pos.\ bias} \\[2pt]
\begin{tabular}{lc}
\toprule
$b_{\mathrm{rel}}$ & Top-1 \\
\midrule
without            & 84.3 \\
\rowcolor{gray!12}with~$\star$       & \textbf{84.4} \\
\bottomrule
\end{tabular}
\end{minipage}
\hfill
\begin{minipage}[t]{0.22\textwidth}
\centering
(d) \textbf{Affinity heads} \\[2pt]
\begin{tabular}{lc}
\toprule
Heads & Top-1 \\
\midrule
\rowcolor{gray!12}1~$\star$ & 84.4 \\
2 & 84.1 \\
4 & \textbf{84.7} \\
\bottomrule
\end{tabular}
\end{minipage}
\end{table}
\vspace{-1mm}

\noindent\textbf{Discussion.} (a)~Pre-SSM placement is the strongest configuration: moving the operator post-SSM costs $0.7$ pt and removing it entirely a further $1.3$ pt, consistent with the three preconditioning pathways identified in Sec.~\ref{sec:preconditioning}. (b)~The default growing-radius $(1,1,2,3)$ occupies the compute--accuracy Pareto front: a globally smaller schedule loses $0.9$ pt and a globally larger one offers only $+0.3$ pt at noticeably higher FLOPs. (c)~The relative-position bias gives a small but consistent gain, indicating feature-conditioned affinities benefit from a light geometric prior. (d)~Multi-head GraphScan is non-monotonic ($H{=}2$ dips below $H{=}1$, $H{=}4$ surpasses both at no added parameter cost); we retain $H{=}1$ for parity with prior work and treat multi-head as a lightweight extension.

\vspace{-2mm}

\section{Conclusion}
\label{sec:concl}

\vspace{-2mm}

We introduced \textbf{GraphScan}, a local windowed graph attention operator that preconditions tokens before the selective state-space scan; in a hierarchical Vision Mamba backbone, this single change yields state-of-the-art accuracy among Vision SSMs on ImageNet-1K, COCO, and ADE20K at comparable parameter and FLOP budgets. Because GraphScan sits on the input side, it is orthogonal to recurrence-side advances~\cite{vssd,spatialmamba,mamba2,mamba3}; composing both directions is a natural next step. To the question posed in the title: yes---graphs do help Vision SSMs see better.

{
\small
\bibliographystyle{ieeenat_fullname}
\bibliography{main}
}

\newpage
\appendix
\section{Analysis of GraphScan-Induced Selective Scan}
\label{app:graphscan_analysis}

This appendix provides a more detailed analysis of how GraphScan modifies the selective scan computation. 
The goal is not to prove a task-level generalization guarantee, but to make precise the architectural effect of routing tokens before the SSM. 
GraphScan changes both the values entering the recurrent state and the token-dependent selective parameters that control propagation and readout.

Throughout this appendix, we omit batch dimensions and use the notation from Sec.~\ref{sec:prelim}. 
All decompositions below are exact unless explicitly stated otherwise.

\subsection{Selective Scan as a Token-Conditioned Dynamical System}
\label{app:ssm_dynamical_system}

For a visual token sequence $\mathbf{X}=[\mathbf{x}_1,\ldots,\mathbf{x}_L]$, the Mamba-style selective scan can be written as a token-conditioned dynamical system:
\begin{equation}
\mathbf{H}_t
=
\bar{\mathbf{A}}_t(\mathbf{x}_t)\odot\mathbf{H}_{t-1}
+
\bar{\mathbf{B}}_t(\mathbf{x}_t)\odot\mathbf{x}_t^{\uparrow_s},
\qquad
\mathbf{y}_t
=
\left\langle
\mathbf{H}_t,
\mathbf{c}_t(\mathbf{x}_t)^{\uparrow_D}
\right\rangle_{N_s}.
\label{eq:app_base_ssm}
\end{equation}
Here $\bar{\mathbf{A}}_t$ and $\bar{\mathbf{B}}_t$ are post-discretization tensors, while $\mathbf{c}_t$ is the input-dependent readout. 
The dependence on $\mathbf{x}_t$ includes the generation of the final positive step size $\boldsymbol{\Delta}_t$, the selective input vector, and the readout vector. 
Thus, unlike a linear time-invariant SSM, the recurrence coefficients are functions of the token sequence \cite{mamba}. 

For a zero initial state, the hidden state unrolls as
\begin{equation}
\mathbf{H}_t
=
\sum_{i=1}^{t}
\mathbf{G}_{t,i}\odot
\bar{\mathbf{B}}_i(\mathbf{x}_i)
\odot
\mathbf{x}_i^{\uparrow_s},
\qquad
\mathbf{G}_{t,i}
=
\bigodot_{j=i+1}^{t}
\bar{\mathbf{A}}_j(\mathbf{x}_j).
\label{eq:app_unroll}
\end{equation}
This expression makes the sequence geometry explicit: information from token $i$ reaches token $t$ through the product of the intermediate state transitions along the scan path.

\subsection{Exact Routed-vs-Base Decomposition}
\label{app:routed_decomposition}

GraphScan replaces each token $\mathbf{x}_t$ with a routed token $\mathbf{x}'_t=\mathbf{x}_t+\mathbf{m}_t$, where, with $M=W_vW_o$ as in Eq.~\eqref{eq:graphscan_preconditioner},
\begin{equation}
\mathbf{m}_t
=
\sum_{j\in\mathcal{S}_r(t)}
\alpha_{tj}\mathbf{x}_jM.
\label{eq:app_routing_message}
\end{equation}
The routed selective parameters are
\[
\bar{\mathbf{A}}'_t=\bar{\mathbf{A}}_t(\mathbf{x}'_t),
\qquad
\bar{\mathbf{B}}'_t=\bar{\mathbf{B}}_t(\mathbf{x}'_t),
\qquad
\mathbf{c}'_t=\mathbf{c}_t(\mathbf{x}'_t),
\]
and the routed recurrence is
\begin{equation}
\mathbf{H}'_t
=
\bar{\mathbf{A}}'_t\odot\mathbf{H}'_{t-1}
+
\bar{\mathbf{B}}'_t\odot(\mathbf{x}'_t)^{\uparrow_s}.
\label{eq:app_routed_ssm}
\end{equation}

\noindent\textbf{Proposition A.1 (Exact routed-vs-base decomposition).}
Let $\delta\mathbf{H}_t=\mathbf{H}'_t-\mathbf{H}_t$ and define
\[
\delta\bar{\mathbf{A}}_t=\bar{\mathbf{A}}'_t-\bar{\mathbf{A}}_t,
\qquad
\delta\bar{\mathbf{B}}_t=\bar{\mathbf{B}}'_t-\bar{\mathbf{B}}_t,
\qquad
\delta\mathbf{c}_t=\mathbf{c}'_t-\mathbf{c}_t.
\]
Assuming the base and routed recurrences share the same initial state, $\delta\mathbf{H}_0=0$, we have
\begin{equation}
\delta\mathbf{H}_t
=
\sum_{i=1}^{t}
\mathbf{G}'_{t,i}\odot
\left[
\bar{\mathbf{B}}'_i\odot\mathbf{m}_i^{\uparrow_s}
+
\delta\bar{\mathbf{B}}_i\odot\mathbf{x}_i^{\uparrow_s}
+
\delta\bar{\mathbf{A}}_i\odot\mathbf{H}_{i-1}
\right],
\label{eq:app_exact_state_decomp}
\end{equation}
where $\mathbf{G}'_{t,i}=\bigodot_{j=i+1}^{t}\bar{\mathbf{A}}'_j$ and $\mathbf{G}'_{t,t}=\mathbf{1}$. 
Moreover,
\begin{equation}
\delta\mathbf{y}_t
=
\left\langle
\delta\mathbf{H}_t,
(\mathbf{c}'_t)^{\uparrow_D}
\right\rangle_{N_s}
+
\left\langle
\mathbf{H}_t,
\delta\mathbf{c}_t^{\uparrow_D}
\right\rangle_{N_s}.
\label{eq:app_exact_output_decomp}
\end{equation}

\noindent\textit{Proof.}
Subtract the base recurrence from the routed recurrence. 
Using $\mathbf{x}'_t=\mathbf{x}_t+\mathbf{m}_t$, $\mathbf{H}'_{t-1}=\mathbf{H}_{t-1}+\delta\mathbf{H}_{t-1}$, $\bar{\mathbf{A}}'_t=\bar{\mathbf{A}}_t+\delta\bar{\mathbf{A}}_t$, and $\bar{\mathbf{B}}'_t=\bar{\mathbf{B}}_t+\delta\bar{\mathbf{B}}_t$, we obtain
\[
\begin{aligned}
\delta\mathbf{H}_t
&=
\bar{\mathbf{A}}'_t\odot(\mathbf{H}_{t-1}+\delta\mathbf{H}_{t-1})
+
\bar{\mathbf{B}}'_t\odot(\mathbf{x}_t+\mathbf{m}_t)^{\uparrow_s}
-
\bar{\mathbf{A}}_t\odot\mathbf{H}_{t-1}
-
\bar{\mathbf{B}}_t\odot\mathbf{x}_t^{\uparrow_s} \\
&=
\bar{\mathbf{A}}'_t\odot\delta\mathbf{H}_{t-1}
+
(\bar{\mathbf{A}}'_t-\bar{\mathbf{A}}_t)\odot\mathbf{H}_{t-1}
+
\bar{\mathbf{B}}'_t\odot\mathbf{m}_t^{\uparrow_s}
+
(\bar{\mathbf{B}}'_t-\bar{\mathbf{B}}_t)\odot\mathbf{x}_t^{\uparrow_s}.
\end{aligned}
\]
Therefore,
\[
\delta\mathbf{H}_t
=
\bar{\mathbf{A}}'_t\odot\delta\mathbf{H}_{t-1}
+
\delta\bar{\mathbf{A}}_t\odot\mathbf{H}_{t-1}
+
\bar{\mathbf{B}}'_t\odot\mathbf{m}_t^{\uparrow_s}
+
\delta\bar{\mathbf{B}}_t\odot\mathbf{x}_t^{\uparrow_s}.
\]
Unrolling this recurrence from $\delta\mathbf{H}_0=0$ gives Eq.~\eqref{eq:app_exact_state_decomp}. 
For the output, write $\mathbf{H}'_t=\mathbf{H}_t+\delta\mathbf{H}_t$ and $\mathbf{c}'_t=\mathbf{c}_t+\delta\mathbf{c}_t$. 
Then
\[
\begin{aligned}
\delta\mathbf{y}_t
&=
\left\langle
\mathbf{H}'_t,
(\mathbf{c}'_t)^{\uparrow_D}
\right\rangle_{N_s}
-
\left\langle
\mathbf{H}_t,
\mathbf{c}_t^{\uparrow_D}
\right\rangle_{N_s} \\
&=
\left\langle
\mathbf{H}_t+\delta\mathbf{H}_t,
(\mathbf{c}'_t)^{\uparrow_D}
\right\rangle_{N_s}
-
\left\langle
\mathbf{H}_t,
\mathbf{c}_t^{\uparrow_D}
\right\rangle_{N_s} \\
&=
\left\langle
\delta\mathbf{H}_t,
(\mathbf{c}'_t)^{\uparrow_D}
\right\rangle_{N_s}
+
\left\langle
\mathbf{H}_t,
(\mathbf{c}'_t-\mathbf{c}_t)^{\uparrow_D}
\right\rangle_{N_s}.
\end{aligned}
\]
This proves Eq.~\eqref{eq:app_exact_output_decomp}. \hfill$\square$

\noindent\textbf{Interpretation.}
Eq.~\eqref{eq:app_exact_state_decomp} separates three hidden-state pathways: direct routed value injection, write modulation, and transition modulation. 
Eq.~\eqref{eq:app_exact_output_decomp} adds a fourth pathway: readout modulation. 
The decomposition is exact and does not rely on treating the GraphScan message as small.

\subsection{Local-Global Kernel Induced by GraphScan}
\label{app:local_global_kernel}

We now isolate the direct routed value path in Eq.~\eqref{eq:app_exact_state_decomp}. 
Define the routed SSM input-to-output operator
\begin{equation}
\mathcal{K}'_{t,i}(\mathbf{z})
=
\left\langle
\mathbf{G}'_{t,i}\odot
\bar{\mathbf{B}}'_i\odot
\mathbf{z}^{\uparrow_s},
(\mathbf{c}'_t)^{\uparrow_D}
\right\rangle_{N_s},
\qquad i\le t.
\label{eq:app_routed_kernel}
\end{equation}
For $i>t$, set $\mathcal{K}'_{t,i}(\mathbf{z})=\mathbf{0}$.

\noindent\textbf{Corollary A.2 (Local-global kernel composition).}
The direct routed value contribution to $\delta\mathbf{y}_t$ is
\begin{equation}
\delta\mathbf{y}^{\mathrm{route}}_t
=
\sum_{j=1}^{L}
\sum_{\substack{i\le t:\\ j\in\mathcal{S}_r(i)}}
\alpha_{ij}\,
\mathcal{K}'_{t,i}(\mathbf{x}_jM).
\label{eq:app_exact_local_global_kernel}
\end{equation}

\noindent\textit{Proof.}
The direct routed value pathway in Eq.~\eqref{eq:app_exact_state_decomp} contributes
\[
\delta\mathbf{y}^{\mathrm{route}}_t
=
\left\langle
\sum_{i=1}^{t}
\mathbf{G}'_{t,i}\odot
\bar{\mathbf{B}}'_i\odot
\mathbf{m}_i^{\uparrow_s},
(\mathbf{c}'_t)^{\uparrow_D}
\right\rangle_{N_s}.
\]
By the definition of $\mathcal{K}'_{t,i}$, this equals
\[
\delta\mathbf{y}^{\mathrm{route}}_t
=
\sum_{i=1}^{t}\mathcal{K}'_{t,i}(\mathbf{m}_i).
\]
Substituting $\mathbf{m}_i=\sum_{j\in\mathcal{S}_r(i)}\alpha_{ij}\mathbf{x}_jM$ gives
\[
\delta\mathbf{y}^{\mathrm{route}}_t
=
\sum_{i=1}^{t}
\mathcal{K}'_{t,i}
\left(
\sum_{j\in\mathcal{S}_r(i)}
\alpha_{ij}\mathbf{x}_jM
\right).
\]
For fixed routed coefficients, $\mathcal{K}'_{t,i}$ is linear in its argument. 
Therefore,
\[
\delta\mathbf{y}^{\mathrm{route}}_t
=
\sum_{i=1}^{t}
\sum_{j\in\mathcal{S}_r(i)}
\alpha_{ij}\,
\mathcal{K}'_{t,i}(\mathbf{x}_jM).
\]
Exchanging the finite sums over $i$ and $j$ gives Eq.~\eqref{eq:app_exact_local_global_kernel}. \hfill$\square$

\noindent\textbf{Interpretation.}
Eq.~\eqref{eq:app_exact_local_global_kernel} shows that GraphScan composes two kernels. 
The first is a local semantic kernel $\alpha_{ij}$ defined over a bounded two-dimensional neighborhood. 
The second is a global selective SSM kernel $\mathcal{K}'_{t,i}$ defined along the one-dimensional scan path. 
Thus, GraphScan does not replace the SSM with local attention; it composes local semantic routing with global state-space propagation.

\subsection{Scan-Distance Shortcuts under Bounded Survival}
\label{app:scan_distance}

The local-global kernel clarifies why routing can help when two-dimensional adjacency is poorly aligned with raster order.

\noindent\textbf{Proposition A.3 (Semantic shortcut under bounded routed survival).}
Assume the routed selective transitions satisfy $\|\bar{\mathbf{A}}'_j\|_{\infty}\le \rho<1$, and assume $\|\bar{\mathbf{B}}'_i\|_{\infty}\le B_{\max}$ and $\|\mathbf{c}'_t\|_{\infty}\le C_{\max}$. 
Then, for any $\mathbf{z}\in\mathbb{R}^{D}$ and any $i\le t$,
\begin{equation}
\|\mathcal{K}'_{t,i}(\mathbf{z})\|_{\infty}
\le
N_s B_{\max}C_{\max}\rho^{t-i}\|\mathbf{z}\|_{\infty}.
\label{eq:app_kernel_bound}
\end{equation}
Consequently, a source token routed through slot $i$ contributes to output $t$ with attenuation governed by the scan distance $t-i$.

\noindent\textit{Proof.}
By definition,
\[
\mathbf{G}'_{t,i}
=
\bigodot_{\ell=i+1}^{t}\bar{\mathbf{A}}'_{\ell}.
\]
Using submultiplicativity of the elementwise $\ell_{\infty}$ norm under Hadamard products and the assumption $\|\bar{\mathbf{A}}'_{\ell}\|_{\infty}\le \rho$, we obtain
\[
\|\mathbf{G}'_{t,i}\|_{\infty}
\le
\prod_{\ell=i+1}^{t}\|\bar{\mathbf{A}}'_{\ell}\|_{\infty}
\le
\rho^{t-i}.
\]
For each output channel $d$, the $d$-th component of $\mathcal{K}'_{t,i}(\mathbf{z})$ is
\[
[\mathcal{K}'_{t,i}(\mathbf{z})]_d
=
\sum_{n=1}^{N_s}
[\mathbf{G}'_{t,i}]_{d,n}
[\bar{\mathbf{B}}'_i]_{d,n}
z_d
[\mathbf{c}'_t]_n.
\]
Taking absolute values and applying the bounds on $\mathbf{G}'_{t,i}$, $\bar{\mathbf{B}}'_i$, $\mathbf{c}'_t$, and $\mathbf{z}$ gives
\[
|[\mathcal{K}'_{t,i}(\mathbf{z})]_d|
\le
\sum_{n=1}^{N_s}
\rho^{t-i}B_{\max}C_{\max}\|\mathbf{z}\|_{\infty}
=
N_sB_{\max}C_{\max}\rho^{t-i}\|\mathbf{z}\|_{\infty}.
\]
Taking the maximum over channels $d$ proves Eq.~\eqref{eq:app_kernel_bound}. \hfill$\square$

\noindent\textbf{Interpretation.}
This result does not claim GraphScan always improves accuracy. 
It shows a structural effect: a source patch can influence the SSM through a routed slot whose scan distance to the output may differ from the source patch's raster distance. 
When two-dimensional adjacency and one-dimensional raster order disagree, GraphScan supplies additional local semantic paths into the selective scan.

\subsection{Reachability View from State-Space Systems}
\label{app:reachability}

If the selective coefficients are frozen, Eq.~\eqref{eq:app_base_ssm} becomes a linear time-varying state-space system. 
Classical control theory studies how inputs enter such systems through transition operators and input matrices, often through reachability or controllability analyses \cite{kalman1960,sontag1998control}. 
GraphScan modifies the input directions before they are injected into the recurrent state. 
In the frozen-coefficient view, the base system is driven by isolated raster tokens, while the routed system is driven by locally aggregated semantic neighborhoods.

More explicitly, with frozen coefficients and zero initial state, the base hidden state is
\begin{equation}
\mathbf{H}_t
=
\sum_{i=1}^{t}
\mathbf{G}_{t,i}\odot
\bar{\mathbf{B}}_i\odot
\mathbf{x}_i^{\uparrow_s}.
\label{eq:app_reach_base}
\end{equation}
With GraphScan-routed inputs but the same frozen coefficients, the hidden state becomes
\begin{equation}
\mathbf{H}^{\mathrm{gs}}_t
=
\sum_{i=1}^{t}
\mathbf{G}_{t,i}\odot
\bar{\mathbf{B}}_i\odot
\left(
\mathbf{x}_i+
\sum_{j\in\mathcal{S}_r(i)}
\alpha_{ij}\mathbf{x}_jM
\right)^{\uparrow_s}.
\label{eq:app_reach_gs}
\end{equation}
Eq.~\eqref{eq:app_reach_gs} shows that GraphScan changes the finite-horizon input-to-state map by introducing cross-token input directions induced by the sparse routing matrix $P(\mathbf{X})$. 
The recurrent state is no longer driven only by isolated raster tokens; it is also driven by local semantic combinations of nearby tokens.

This reachability view should be interpreted carefully: GraphScan does not guarantee better controllability or generalization in the formal control-theoretic sense without additional assumptions. 
Instead, it provides a structural mechanism by which the finite recurrent state receives richer local semantic inputs before global propagation.

\subsection{Baseline Containment}
\label{app:containment}

\noindent\textbf{Proposition A.4 (Baseline containment).}
GraphScan-Mamba contains the corresponding base Vision Mamba model as a subnetwork.

\noindent\textit{Proof.}
From Eq.~\eqref{eq:graphscan_update}, setting $W_o=0$ makes the routed message vanish for every token, so $\mathbf{x}'_i=\mathbf{x}_i$. 
If an implementation includes a bias in the output projection, setting that bias to zero as well gives the same identity map. 
Equivalently, Eq.~\eqref{eq:graphscan_preconditioner} reduces to $\mathbf{X}'=\mathbf{X}$. 
The subsequent selective SSM, residual branches, feed-forward layers, and classifier are then identical to those of the corresponding base Vision Mamba model. 
Therefore, the base model is obtained as a parameter setting of GraphScan-Mamba. \hfill$\square$

\noindent\textbf{Interpretation.}
This containment result does not imply optimization will always find a better solution, but it shows that GraphScan preserves the base model as a special case while adding trainable local routing paths before the SSM.

\subsection{Relation to Local Attention and Vision GNNs}
\label{app:relation_local_attention_gnn}

GraphScan uses a primitive related to local neighborhood attention and graph message passing.
Neighborhood attention restricts Q/K/V attention to a sliding neighborhood around each token, providing an efficient local attention mechanism for hierarchical Transformer backbones \cite{nat}; non-local networks generalize visual feature aggregation through pairwise interactions \cite{nonlocal}.
Vision GNNs instead view image patches as graph nodes and perform learned neighborhood aggregation over visual tokens \cite{vig,pvg}, building on the broader graph-network and message-passing primitives studied in deep learning \cite{battaglia2018relational,gilmer2017mpnn}.
GraphScan borrows this local affinity aggregation primitive, but uses it in a different role: it is not the main token mixer of a Transformer or GNN backbone.
It is a scan-time preconditioner placed immediately before the selective SSM.
As shown above, this placement causes local semantic routing to influence both the values entering the recurrent state and the token-conditioned parameters of the selective scan.

\section{Extended Related Work}
\label{app:related_extended}

This appendix expands the condensed related work in Sec.~\ref{sec:related} with the fuller treatment across all four prior-work paradigms.

\subsection{Vision Backbones}
\label{app:rw_backbones}

Visual backbone design has evolved through several representation paradigms.
Convolutional neural networks model images as regular grids and encode locality through sliding filters, hierarchical downsampling, residual connections, scaling rules, and large-kernel convolutional blocks \cite{lenet,alexnet,vgg,googlenet,resnet,densenet,efficientnet,regnet,convnext,convnextv2,slak,unireplknet,inceptionnext}.
Vision Transformers recast images as sequences of patch tokens and use self-attention for flexible long-range interaction \cite{transformer,vit,deit}.
To support high-resolution recognition and dense prediction, later Transformer backbones introduced hierarchical designs, shifted or windowed attention, pyramid structures, cross-shaped and focal attention, neighborhood attention, deformable attention, bi-level routing, non-local interactions, and convolution-attention hybrids \cite{swin,pvt,pvtv2,twins,cswin,focal,nat,dat,nonlocal,biformer,maxvit,coatnet,vitadapter,internimage}.
Beyond convolution and attention, all-MLP models explore token mixing with feed-forward layers \cite{mlpmixer,resmlp,gmlp}, while Vision GNNs view image patches as graph nodes and propagate information through learned neighborhood aggregation \cite{gcn,gat,battaglia2018relational,vig,pvg}.

GraphScan is related to this graph-based view, but it is not a standalone Vision GNN backbone.
Instead, it uses lightweight local graph routing inside a Vision SSM, enriching visual tokens with semantic neighborhood context before global state-space mixing.

\subsection{State Space Sequence Models}
\label{app:rw_ssm}

Structured state space models provide efficient sequence modeling with long-range interactions and linear or near-linear scaling in sequence length \cite{s4,s4d,s5}.
Early SSM-based architectures such as H3 and Hyena demonstrated the promise of long convolutional and state-space sequence mixers as alternatives to attention \cite{h3,hyena}.
Mamba introduced selective state spaces, making key SSM parameters input-dependent so that the model can selectively propagate, reset, or ignore information along the sequence while retaining efficient scan-based computation \cite{mamba}.
Mamba-2 further connected SSMs and attention through structured state space duality and improved the efficiency of the core sequence mixer \cite{mamba2}.
Mamba-3 extends this line with recurrence-level improvements such as exponential-trapezoidal discretization, complex-valued state transitions, and MIMO updates \cite{mamba3}.

These works improve the sequence model itself.
Our work addresses a complementary question that arises when such models are used for vision: how should a two-dimensional visual feature map be routed before it is processed by a one-dimensional state-space operator?

\subsection{Vision Mamba and Scan Design}
\label{app:rw_vssm_scans}

Vision Mamba models adapt selective SSMs to images by converting feature maps into token sequences.
Vim applies bidirectional Mamba blocks to visual sequences, while VMamba introduces a two-dimensional selective scan to process visual features along multiple spatial directions \cite{vim,vmamba}.
Since scan order determines how two-dimensional structure is exposed to the one-dimensional recurrence, subsequent work has explored increasingly structured scan patterns.
Continuous scans improve spatial continuity \cite{plainmamba}; local-window scans preserve short-range dependencies and search layer-wise scan choices \cite{localmamba}; atrous and multi-scale scans improve efficiency and hierarchical coverage \cite{efficientvmamba,msvmamba}; grouped and multidimensional scans distribute scanning across channels or spatial axes \cite{groupmamba,mamband}; and fractal or space-filling scans aim to preserve locality under serialization \cite{fractalmamba}.

These methods show that scan design is central to Vision SSM performance.
However, most of them rely on predefined traversal patterns or layer-level choices, rather than input-dependent semantic routing.

\subsection{Dynamic and Relational Scanning}
\label{app:rw_dynamic_scans}

Recent work investigates adaptive scanning for Vision SSMs.
Quadtree-based scan methods learn spatial partitions with varying granularity \cite{quadmamba}.
Coordinate-offset methods predict deformable sampling locations and construct routed tokens through bilinear interpolation before feeding them to the SSM \cite{damamba,defmamba}.
Other approaches reorder or prioritize tokens using task- or content-dependent scores, or use similarity signals in specialized correspondence settings \cite{objectness_scan,asm_unet,mambamatcher}.
PVMamba modifies hidden-state generation through dynamic state aggregation and spatial sampling to reduce sequential constraints \cite{pvmamba}.
Spatial-Mamba introduces structure-aware fusion in the state space \cite{spatialmamba}, VSSD reformulates SSD into a non-causal visual operator \cite{vssd}, and MambaVision hybridizes Mamba and Transformer blocks for high-level visual modeling \cite{mambavision}.

GraphScan occupies a distinct design space.
Rather than changing the SSM recurrence, collapsing scan order, or predicting continuous coordinates, GraphScan learns a local semantic graph over visual tokens and performs one-step message passing before the SSM.
It is input-adaptive like dynamic scans, spatially bounded like local scans, and semantic like graph message passing.
Consequently, the global selective SSM receives tokens that are already locally grounded, while preserving token count and linear scaling.

\section{Architecture Variants}
\label{app:variants}

Table~\ref{tab:variants} reports per-variant channel widths, stage depths, MLP ratios, graph radii, parameters, and FLOPs for the three GraphScan-Mamba scales used throughout the experiments.

\begin{table}[h]
\centering
\setlength{\tabcolsep}{4pt}
\caption{\textbf{GraphScan-Mamba variants.} Channels $(C_1,C_2,C_3,C_4)$, depths $(N_1,N_2,N_3,N_4)$, MLP ratios $(m_1,m_2,m_3,m_4)$, and graph radii $(r_1,r_2,r_3,r_4)$. Parameters and FLOPs reported at $224{\times}224$ input.}
\label{tab:variants}
\footnotesize
\begin{tabular}{lcccccc}
\toprule
Variant & Channels & Depths & MLP ratios & Radii & Params & FLOPs \\
\midrule
GraphScan-Mamba-T & 80/160/320/512  & 3/4/12/5 & 4/4/3/3 & 1/1/2/3 & 28M & 5.2G  \\
GraphScan-Mamba-S & 96/192/384/512  & 4/8/20/6 & 4/4/3/3 & 1/1/2/3 & 49M & 11.1G \\
GraphScan-Mamba-B & 112/224/448/640 & 4/8/25/8 & 4/4/3/4 & 1/1/2/3 & 93M & 17.8G \\
\bottomrule
\end{tabular}
\end{table}

\section{Visualizations}
\label{app:visualizations}

\begin{figure}[h]
\centering
\includegraphics[width=\textwidth]{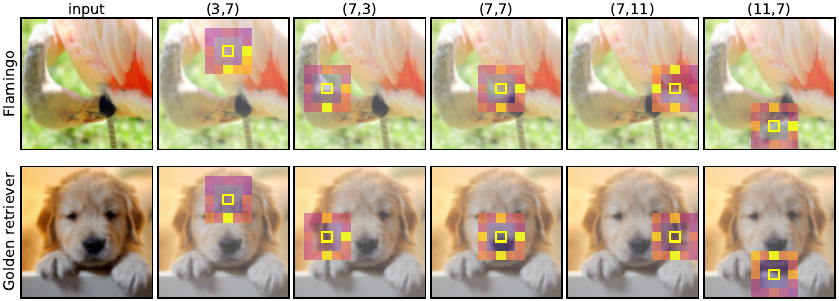}
\caption{\textbf{Stage-3 GraphScan attention} on two ImageNet val images. For each query patch (yellow box), the $14{\times}14$ attention heatmap concentrates on the $L_{\infty}{=}2$ outer ring; the central $3{\times}3$ block is suppressed because the preceding depthwise convolution already covers it.}
\label{fig:stage3_attn}
\end{figure}

\begin{figure}[h]
\centering
\includegraphics[width=\textwidth]{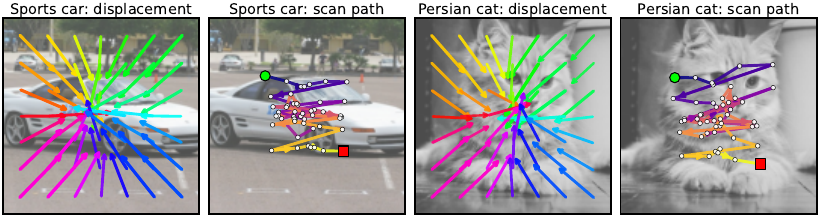}
\caption{\textbf{Stage-4 effective routing.} Per-patch displacement field (HSV by direction) and the polyline visiting displaced positions in raster order (plasma, start \textcolor{green!50!black}{$\bullet$}, end \textcolor{red}{$\blacksquare$}). Both converge onto the discriminative object content.}
\label{fig:stage4_routing}
\end{figure}

Capturing the per-block attention weights of the trained GraphScan-Mamba-S backbone surfaces two patterns. (i)~\emph{Stage-3 outer-ring specialization} (Fig.~\ref{fig:stage3_attn}): the depthwise $3{\times}3$ convolution that runs before GraphScan already aggregates the inner $3{\times}3$ block into the query, so the operator suppresses the inner ring by roughly $40\times$ below uniform and attends almost exclusively to the $L_{\infty}{=}2$ outer ring---specializing in cells the preceding convolution cannot reach. (ii)~\emph{Stage-4 convergent global routing} (Fig.~\ref{fig:stage4_routing}): with the local window covering the full $7{\times}7$ feature map, the displacement field $\Delta\widehat{\mathbf{p}}_i = \sum_j \alpha_{ij}(\mathbf{p}_j - \mathbf{p}_i)$ converges onto salient object content and the effective scan path collapses there rather than tracing the spatial grid.

\end{document}